# SE-LIO: Semantics-enhanced Solid-State-LiDAR-Inertial Odometry for Tree-rich Environments

Tisheng Zhang, Linfu Wei, Hailiang Tang, Liqiang Wang, Man Yuan, and Xiaoji Niu

*Abstract*—In this letter, we propose a semantics-enhanced solid-state-LiDAR-inertial odometry (SE-LIO) in tree-rich environments. Multiple LiDAR frames are first merged and compensated with the inertial navigation system (INS) to increase the point-cloud coverage, thus improving the accuracy of semantic segmentation. The unstructured point clouds, such as tree leaves and dynamic objects, are then removed with the semantic information. Furthermore, the pole-like point clouds, primarily tree trunks, are modeled as cylinders to improve positioning accuracy. An adaptive piecewise cylinder-fitting method is proposed to accommodate environments with a high prevalence of curved tree trunks. Finally, the iterated error-state Kalman filter (IESKF) is employed for state estimation. Point-to-cylinder and point-to-plane constraints are tightly coupled with the prior constraints provided by the INS to obtain the maximum a posteriori estimation. Targeted experiments are conducted in complex campus and park environments to evaluate the performance of SE-LIO. The proposed methods, including removing the unstructured point clouds and the adaptive cylinder fitting, yield improved accuracy. Specifically, the positioning accuracy of the proposed SE-LIO is improved by 43.1% compared to the plane-based LIO.

## I. INTRODUCTION

CONTINUOUS, reliable, and accurate positioning in complex environments is crucial for autonomous vehicles and mobile robots. While LiDAR-inertial odometry (LIO) based on line and plane features has demonstrated excellent performance in structured environments, it encounters significant challenges in scenes abundant with unstructured features, such as tree-rich campuses and parks. Fig. 1 illustrates some typical campus and park scenes. In these scenarios, roads are surrounded by trees, resulting in a high proportion of unstructured point clouds (mainly tree leaves point clouds) in the LiDAR-scanned point cloud. These unstructured point clouds may significantly decrease the accuracy of traditional positioning methods that rely on geometric features, such as extracting plane feature points from unstructured point clouds like tree leaves. These plane feature points lack sufficient accuracy, leading to reduced positioning accuracy.

Normal distribution transform (NDT)-based methods [1], [2] have been presented to address these problems. These methods rely on statistical features of point clouds, such as the mean and covariance, rather than geometric features. They have demonstrated commendable positioning accuracy in unstructured environments. However, the matching accuracy and efficiency of NDT-based methods are intrinsically linked to the size of the grid into which they are divided. A smaller grid size yields higher matching accuracy but compromises matching efficiency, and vice versa. Adaptive voxel mapping [3] is a similar method. It divides point clouds into voxel grids, each containing an octree. The octree nodes contain the distribution information of the point clouds in the node. If the point clouds in the node do not conform to a planar distribution, the node is divided into eight sub-nodes, and the distribution of the point clouds is further determined in the sub-nodes, achieving adaptive planar fitting.

Additional methods preprocess point clouds to mitigate the impact of unstructured point clouds. For instance, LeGO-LOAM [4] segregates point clouds into ground and non-ground points, further segmenting the non-ground points and filtering out small non-ground point cloud clusters. It effectively reduces the influence of unstructured point clouds. The segmentation method employed is based on traditional image processing methods [5] for range images. Recently, many networks have been developed for semantic segmentation of point clouds [6]. These models can be categorized into range image-based models [7], [8], voxel-based models [9], [10], and point-based models [11], [12] based on the segmentation objects. As research progresses, the point cloud semantic segmentation accuracy has improved, effectively distinguishing between different types of point clouds, such as ground, buildings, tree trunks, tree leaves, and vehicles. This semantic information can be used to enhance positioning. Several studies have applied semantic segmentation to LiDAR positioning. SuMa++ [13] uses semantic segmentation results to weight feature points, reducing the impact of dynamic objects. PSF-LO [14] performs geometric modeling on several types of semantic segmentation

This research is funded by the National Natural Science Foundation of China (No.42374034, No.41974024) and the National Key Research and Development Program of China (No. 2020YFB0505803). (*Corresponding authors: Hailiang Tang; Xiaoji Niu.*)

Tisheng Zhang, Linfu Wei, Hailiang Tang, Liqiang Wang, Man Yuan and Xiaoji Niu are with the GNSS Research Center, Wuhan University, Wuhan 430079, China. (email: zts@whu.edu.cn; weilf@whu.edu.cn; thl@whu.edu.cn; wlq@whu.edu.cn; yuanman@whu.edu.cn; xjniu@whu.edu.cn)

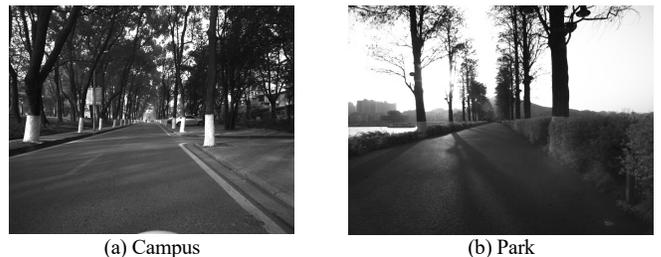

(a) Campus        (b) Park
Fig. 1. Typical scenes abundant with unstructured features.

results types, modeling roads, buildings, and traffic signs as planar features and poles as line features. SLOAM [15] is designed for tree-rich scenes, modeling tree trunks as cylindrical features, but it does not consider the curvature of tree trunks and is therefore not suitable for curved tree-rich environments. SA-LOAM [16] uses semantic information for front-end odometry and loop detection, reducing incorrect matches using semantic label constraints. However, the research mentioned above, which utilizes semantic information to enhance positioning, is mainly designed for spinning LiDAR. It involves projecting the point clouds into a range image for semantic segmentation, which is unsuitable for solid-state LiDAR.

Recently, the low-cost solid-state LiDARs have been widely used in autonomous robots [17]. The solid-state LiDAR has an irregular scanning pattern, completely different from the spinning LiDAR. Consequently, as the scanning time increases, the point-cloud coverage gradually increases, known as the integral property of solid-state LiDAR. Point clouds scanned by solid-state LiDAR are not arranged in a regular array and cannot be directly converted into a range image for semantic segmentation. Hence, a point-based semantic segmentation model would be more appropriate than a range image-based model. However, the point cloud obtained by a single frame (generally 0.1 seconds) of solid-state LiDAR is relatively sparse and low point-cloud coverage [18]. This sparsity often results in missed point clouds from many objects, decreasing semantic segmentation accuracy. While the point-cloud coverage can be increased by extending the scanning cycle of a single-frame point cloud when the carrier is stationary, motion distortion occurs in the point cloud when the carrier is in motion. The longer the scanning cycle, the more notable the effect of motion distortion, which is detrimental to semantic segmentation. With the development of multi-source fusion technology, IMU has become a standard equipment for LiDAR positioning and has been used to compensate for motion distortion [19]. Furthermore, IMU information is tightly coupled with LiDAR information to improve positioning accuracy and robustness [20]–[22]. Therefore, IMU information can be used to compensate for motion distortion, improving the point-cloud coverage and, consequently, the semantic segmentation accuracy.

We propose a semantics-enhanced solid-state-LiDAR-inertial odometry (SE-LIO) for scenes abundant with unstructured features. The proposed method leverages an inertial navigation system (INS) pose to merge and compensate for multiple LiDAR frames. A deep-learning model is employed for semantic segmentation. The segmentation results are then utilized to remove unstructured point clouds and incorporate cylindrical features into state estimation, enhancing positioning accuracy. The primary contributions are as follows:

● To address the low semantic segmentation accuracy caused by sparse point clouds of the solid-state LiDAR, we design an INS-enhanced semantic segmentation method, which leverages the INS pose to merge and compensate for multiple LiDAR frames, thereby improving the point-cloud coverage and semantic segmentation accuracy.

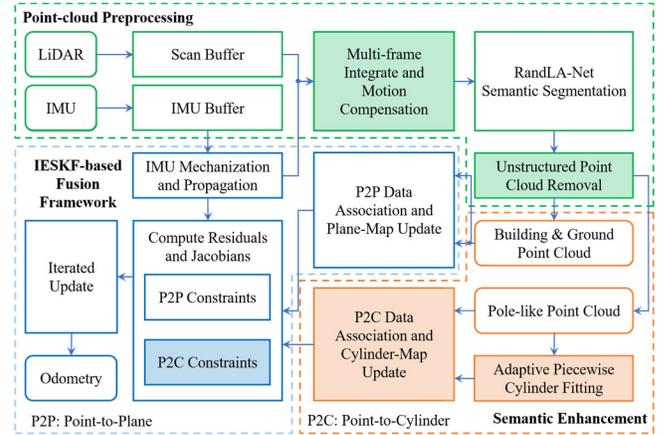

Fig. 2. System overview of the proposed SE-LIO.

● To fully leverage pole-like semantic information, we propose an adaptive piecewise cylinder fitting method, effectively accommodating environments with curved trees, thereby enhancing the system's environmental adaptability and positioning accuracy.

● Real-world experiments were conducted in complex campus and park environments to verify the accuracy and robustness of the proposed method. Several alation experiments were carried out to fully evaluate the impacts of the factors that may influence the accuracy of the proposed SE-LIO.

The remainder of this paper is organized as follows. We give an overview of the system pipeline in Section II. The proposed method is presented in Section III. The experiments and results for quantitative evaluation are discussed in Section IV. Finally, we conclude the proposed method.

## II. SYSTEM OVERVIEW

The system workflow is illustrated in Fig. 2. The point clouds and IMU data are first accumulated until a certain threshold is reached. The current pose is then propagated forward using the IMU mechanization, and the point clouds undergo motion compensation. Following this, the motion-compensated point clouds are subjected to semantic segmentation, dividing them into different types: ground, pole-like, building, tree leaves, and dynamic objects. The unstructured point clouds, primarily tree leaves and dynamic objects, are removed to mitigate their impacts on positioning accuracy. Subsequently, the pole-like semantic information is leveraged to enhance positioning accuracy, including adaptive piecewise cylinder fitting of pole-like point clouds and data association. Finally, the iterated error-state Kalman filter (IESKF) is employed for state estimation. Cylindrical features and plane features are used to construct point-to-cylinder and point-to-plane constraints. These constraints are tightly coupled with the prior constraints provided by INS to obtain the maximum a posteriori estimation.

## III. METHODOLOGY

This section introduces the methodology of SE-LIO. It begins with the point cloud preprocessing phase, followed by the semantic enhancement phase for pole-like point clouds, focusing on adaptive piecewise cylinder fitting and data association methods. Finally, it presents the state estimation algorithm.

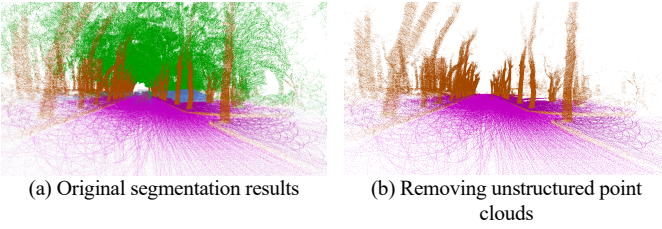

(a) Original segmentation results    (b) Removing unstructured point clouds

Fig. 3. Semantic segmentation results. The purple, green, and brown points represent ground, tree leaves, and pole-like point clouds, respectively.

### A. Point-cloud Preprocessing

The adopted LiDAR is a solid-state non-repetitive scanning LiDAR. Unlike traditional spinning LiDAR, the solid-state-LiDAR has an irregular scanning pattern, and the point clouds of a single frame are relatively sparse. Consequently, range image-based semantic segmentation methods are unsuitable. Instead, we employ a point-based semantic segmentation method, RandLA-Net [12]. Given the relative sparsity of the point cloud from a single frame, we leverage the INS pose to merge and compensate for multiple LiDAR frames, thereby improving the point-cloud coverage and semantic segmentation accuracy.

Specifically, a fixed-length buffer is established to cache point cloud and IMU measurement. When the number of point clouds in the buffer reaches a predetermined threshold, the point clouds are projected to the end of the last frame of the point cloud in the buffer using INS pose, thereby generating the motion-compensated point clouds.

The motion-compensated point clouds are fed into the RandLA-Net [12] model for semantic segmentation. However, the pre-trained RandLA-Net is only suitable for spinning LiDAR and not solid-state LiDAR. Therefore, we performed transfer learning to adapt RandLA-Net to solid-state LiDAR. The segmentation result is depicted in Fig. 3(a), where the purple, green, and brown points represent ground, tree leaves, and pole-like point clouds, respectively. Owing to the substantial influence of unstructured point clouds on positioning accuracy, tree leaves, and dynamic objects point clouds are removed, as depicted in Fig. 3(b).

### B. Semantic Enhancement for Pole-like Point Clouds

After removing unstructured point clouds, the remaining point clouds predominantly consist of ground, building, and pole-like point clouds, such as tree trunks. We employ the cylinder model to fit these pole-like point clouds. Considering the curvature of pole-like objects in the environment, a piecewise fitting method is utilized to adaptively fit these curved pole-like point clouds. The specific method is described in detail below. First, the method for fitting a single cylinder is introduced, followed by the segmented fitting method for fitting a single pole-like object. Subsequently, the method for updating the cylinder map formed by multiple pole-like objects is introduced. Finally, the method for data association is introduced.

#### 1) Single Cylinder Fitting

A cylinder can typically be represented by a minimum of five parameters, with four parameters delineating the axis and one parameter indicating the radius. However, this minimal parameter representation lacks intuitiveness. Therefore, we select a seven-parameter representation for the cylinder model, as shown in the following equation.

$$c = (u^T, q^T, r)^T, \tag{1}$$

where $u$ represents the unit vector in the direction of the axis, $q$ represents a point on the axis, and $r$ represents the radius of the cylinder. Given a cluster of pole-like point clouds obtained by semantic segmentation, let $P = \{p\}$ represent this cluster. The procedure of fitting a cluster of pole-like point clouds to a cylinder can be summarized in the following steps:

*Step 1*: Estimate the axis direction $u$. It is achieved by solving for the maximum eigenvalue of the point cloud covariance. The corresponding eigenvector is the estimated axis vector $u$.

*Step 2*: Construct the rotation matrix $R$ using the axis $u$ to transform the point cloud $P$, *i.e.*, $p' = Rp$, such that the distribution of the transformed point cloud in the z-axis is maximized. The point is then projected onto the x-O-y plane, converting the three-dimensional cylinder fitting problem into a two-dimensional circle fitting problem.

*Step 3*: Employ the random sample consensus (RANSAC) algorithm [23] to calculate the circle parameters. Within RANSAC, the least squares method fits the circle to minimize the sum of the distances between all given points and the circle. The circle equation is expressed as: $(x - x_0)^2 + (y - y_0)^2 = r^2$, where $x_0$ and $y_0$ represent the two-dimensional coordinates of the circle center, and $r$ represents the radius.

*Step 4*: Convert the circle parameters to cylinder parameters. The rotation matrix $R$ is used to convert the circle center coordinates $(x_0, y_0)$ to the point $q$ on the cylinder axis, and the radius $r$ is used as the cylinder radius. Finally, the parameters $c = (u^T, q^T, r)^T$ of the cylinder are obtained.

#### 2) Adaptive Piecewise Cylinder Fitting for Pole-like Point Clouds

Given the diverse curvature of pole-like objects, a single-cylinder model may not provide an optimal fit. Consequently, we propose an adaptive piecewise fitting approach for pole-like objects, as shown in Algorithm 1. A binary tree structure manages the cylinders associated with the same pole-like object, facilitating piecewise fitting.

In Algorithm 1, the pole-like point cloud is first fitted to a cylinder. If the fitting residual is less than the threshold, the pole-like object is deemed to have been fitted. Otherwise, the pole-like point cloud is divided into upper and lower parts, each fitted to a cylinder. This process is repeated until the fitting residual is less than the threshold or the maximum depth of the binary tree is reached.

#### 3) Cylinder Map Update

Given limited pole-like objects, the efficiency of the algorithm will not be significantly impacted even if a linear search is employed for matching. Therefore, we use a linear array to store all trees. For a new frame of pole-like point

---

**Algorithm 1**: Adaptive Piecewise Cylinder Fitting

---
**Input**: Point cloud $P$, depth $d$.
**Output**: Tree node $n$.
**procedure** BUILD_TREE($P$, $d$)
    **if** $d > D_{max}$ **then** // $D_{max}$ represents the maximum depth of the binary tree
        **return** None
    **end**
    $C \leftarrow$ FIT_CYLINDER($P$)
    **if** $\varepsilon . \varepsilon < \overline{\varepsilon}_{max}$ **then** // $\overline{\varepsilon}_{max}$ represents the maximum fitting error
        **return** $C$
    **else**
        // Divide the point cloud into upper and lower parts
        $P_u$, $P_d \leftarrow$ DIVIDE_POINT_CLOUD($P$) // $P_u$ represents the upper part, and $P_d$ represents the lower part
        $C_u \leftarrow$ BUILD_TREE($P_u$, $d + 1$) // $C_u$ represents the upper child
        $C_d \leftarrow$ BUILD_TREE($P_d$, $d + 1$) // $C_d$ represents the lower child
    **end**
**end**

---

**Algorithm 2**: Cylinder Map Update

---
**Input**: Point cloud $P$, Cylinder map $M$.
**Output**: Updated Cylinder map $M$.
**procedure** UPDATE_MAP($P$, $M$)
    **if** not initialized **then**
        add $P$ to buffer and return **if** buffer is not full
        initialized $\leftarrow$ true
    add $P$ to buffer
    **for** each $p$ in $P$ **do**
        **if** $p$ does not belong to any tree **then**
            **if** enough nearest points around $p$ **then**
                create new tree from cluster and add to $M$
            **end**
        **else**
            mark $p$ as a point to update
        **end**
    **end**
    UPDATE_TREES_IN_MAP($M$)
    DELETE_OLD_POINTS_IN_BUFFER()
**end**

---

**Algorithm 3**: Point-to-Cylinder Data Association

---
**Input**: point $p$, cylinder map $M$.
**Output**: cylinder $C$ that point $p$ belongs to.
**procedure** ASSOCIATE_POINT_TO_CYLINDER($p$, $M$)
    // Coarse matching: find the nearest tree to the point $p$
    $T_{near} \leftarrow$ None // $T_{near}$ represents the nearest tree
    $d_{min} \leftarrow \infty$ // $d_{min}$ represents the minimum distance
    **for** tree in $M$ **do**
        distance $\leftarrow$ COMPUTE_DISTANCE_TO_TREE($p$, tree)
        **if** distance $< d_{min}$ **then**
            $T_{near} \leftarrow$ tree
            $d_{min} \leftarrow$ distance
        **end**
    **end**
    // Fine matching: find the cylinder that point $p$ belongs to in the nearest tree
    $C \leftarrow$ None
    **if** $d_{min} <$ threshold **then**
        $C \leftarrow$ FIND_CYLINDER_IN_TREE($p$, $T_{near}$)
    **end**
    **return** $C$
**end**

---

clouds, a coarse matching method is employed to match it with the pole-like objects in the map. If the match is successful, the pole-like object in the map is updated; otherwise, the point does not belong to any tree in the map. We employ the density-based spatial clustering of applications with noise (DBSCAN) [24] algorithm to cluster these new points. The map-update algorithm is shown in Algorithm 2. It

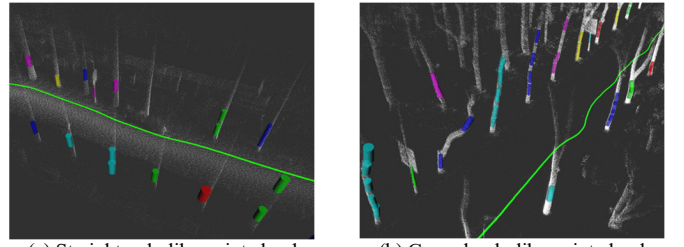

(a) Straight pole-like point cloud    (b) Curved pole-like point cloud
Fig. 4. Cylinder fitting results.

was employed to fit cylinders and update the cylinder map using the campus dataset, yielding the results shown in Fig. 4. As shown in Fig. 4, the proposed method can fit cylinders to straight and curved pole-like objects. For curved pole-like objects, the proposed method can adaptively fit them in segments, rather than using a single-cylinder model.

### 4) Point-to-Cylinder Data Association

The matching process employs a coarse-to-fine matching strategy, as shown in Algorithm 3. The nearest tree to the current point is identified via a linear nearest neighbor search. If the distance between the current point and the nearest tree is less than the predetermined threshold, the point is considered part of that tree. Conversely, if the distance exceeds the threshold, the point is deemed independent of any tree. After confirming the tree to which the point belongs, the specific cylinder to which the point belongs on the tree is determined using binary search.

### C. State Estimation

#### 1) State Definition and Propagation

We employ the IESKF framework to fuse point-to-plane and point-to-cylinder observations. The state vector is defined as

$$\boldsymbol{x} = [(\mathbf{R}_b^w)^T \quad (\boldsymbol{p}^w)^T \quad (\boldsymbol{v}^w)^T \quad \boldsymbol{b}_a^T \quad \boldsymbol{b}_g^T]^T \in SO(3) \times \mathbb{R}^{12}, \quad (2)$$

where $\mathbf{R}_b^w$ represents the rotation matrix of the IMU frame ($b$-frame) relative to the world frame ($w$-frame), $\boldsymbol{p}^w$ and $\boldsymbol{v}^w$ represents the translation and velocity vector of the $b$-frame relative to the $w$-frame, respectively. $\boldsymbol{b}_a$ and $\boldsymbol{b}_g$ represents the accelerometer and gyroscope biases, respectively. The estimated state and error state are defined as

$$\hat{\boldsymbol{x}} = [(\hat{\mathbf{R}}_b^w)^T \quad (\hat{\boldsymbol{p}}^w)^T \quad (\hat{\boldsymbol{v}}^w)^T \quad \hat{\boldsymbol{b}}_a^T \quad \hat{\boldsymbol{b}}_g^T]^T \in SO(3) \times \mathbb{R}^{12}$$
$$\delta\boldsymbol{x} = \boldsymbol{x} \boxminus \hat{\boldsymbol{x}} = [\delta\boldsymbol{\theta}^T \quad \delta\boldsymbol{p}^T \quad \delta\boldsymbol{v}^T \quad \delta\boldsymbol{b}_a^T \quad \delta\boldsymbol{b}_g^T]^T \in \mathbb{R}^{15} \quad (3)$$

Before the observation update, the error covariance matrix is propagated using the motion model. The employed motion model is similar to that of FAST-LIO [21], except that we use a first-order Gaussian-Markov model [25] to model IMU bias, whereas FAST-LIO uses a random walk model to model IMU bias. The discrete IMU bias model employed is defined as

$$\delta\boldsymbol{b}_{a,k+1} = \left(1 - \frac{1}{T_a}\Delta t\right)\delta\boldsymbol{b}_{a,k} + \boldsymbol{n}_a$$
$$\delta\boldsymbol{b}_{g,k+1} = \left(1 - \frac{1}{T_g}\Delta t\right)\delta\boldsymbol{b}_{g,k} + \boldsymbol{n}_g \quad (4)$$

where $T_a$ and $T_g$ represent the correlation time of the accelerometer bias and gyroscope bias, respectively, and $\boldsymbol{n}_a$ and $\boldsymbol{n}_g$ represent the Gaussian white noise of the accelerometer bias and gyroscope bias, respectively.

The other error state transition equations of $\delta\boldsymbol{x} = \boldsymbol{x} \boxminus \hat{\boldsymbol{x}}$ are as follows

$$
\begin{aligned}
\delta\boldsymbol{\theta}_{k+1} &= \mathrm{Exp}(-(\boldsymbol{\omega}^b - \boldsymbol{b}_g)\Delta t)\delta\boldsymbol{\theta}_k - \delta\boldsymbol{b}_g\Delta t - \boldsymbol{n}_\theta \\
\delta\boldsymbol{p}_{k+1} &= \delta\boldsymbol{p}_k + \delta\boldsymbol{v}\Delta t \\
\delta\boldsymbol{v}_{k+1} &= \delta\boldsymbol{v}_k - \hat{\mathbf{R}}_b^w\Delta t\delta\boldsymbol{b}_a - \hat{\mathbf{R}}_b^w((\boldsymbol{f}^b - \boldsymbol{b}_a)\times)\Delta t\delta\boldsymbol{\theta}_k \\
&\quad - \boldsymbol{n}_v
\end{aligned}
\tag{5}
$$

where Exp represents the exponential mapping from Lie algebra to Lie group, $\boldsymbol{n}_\theta$ and $\boldsymbol{n}_v$ represent the Gaussian white noise of the attitude and velocity, $(\cdot\times)$ represents the conversion of a vector to a skew-symmetric matrix.

### 2) Point-to-Cylinder Measurement Model

The point observed by the LiDAR is represented as $\boldsymbol{p}_i^l$, where $i$ represents the $i$-th point, and $l$ represents the LiDAR frame ($l$-frame). Assuming the LiDAR-IMU extrinsic parameters have been calibrated, the $i$-th point $\boldsymbol{p}_i^b$ in $b$-frame can be obtained by transforming the point $\boldsymbol{p}_i^l$, *i.e.* $\boldsymbol{p}_i^b = \mathbf{R}_l^b(\boldsymbol{p}_i^l - \boldsymbol{p}_l^b)$. Here, $\boldsymbol{p}_l^b$ represents the translation vector of the $l$-frame relative to the $b$-frame, and $\mathbf{R}_l^b$ represents the rotation matrix of the $l$-frame relative to the $b$-frame.

Given the point cloud set $P_c = \{\boldsymbol{p}_i^b\}$ of the pole-like object obtained from semantic segmentation, where $i$ represents the $i$-th point, the cylinder $\boldsymbol{c} = (\boldsymbol{u}^T, \boldsymbol{q}^T, r)^T$ to which the point $\boldsymbol{p}_i^b$ belongs can be determined using the data association method described in Section III-B. The distance from the point to the cylinder surface is defined as

$$
d_{\boldsymbol{c},\boldsymbol{p}_i^b} = ||(\boldsymbol{u}\times)(\mathbf{R}_b^w\boldsymbol{p}_i^b + \boldsymbol{p}^w - \boldsymbol{q})||_2. \tag{6}
$$

The residual of $\boldsymbol{p}_i^b$ to the cylinder can be written as

$$
\begin{aligned}
h_c(\boldsymbol{x}_k, \boldsymbol{p}_i^b) &= d_{\boldsymbol{c},\boldsymbol{p}_i^b} \\
&= h_c(\hat{\boldsymbol{x}}_k \boxplus \delta\boldsymbol{x}_k, \boldsymbol{p}_i^b) + \boldsymbol{n}_i \\
&\approx h_c(\hat{\boldsymbol{x}}_k, \boldsymbol{p}_i^b) + \mathbf{H}_{c,i}(\hat{\boldsymbol{x}}_k, \boldsymbol{p}_i^b)\delta\boldsymbol{x}_k + \boldsymbol{n}_i
\end{aligned}
\tag{7}
$$

where $\boldsymbol{n}_i$ represents the Gaussian white noise, and $\mathbf{H}_{c,i}$ represents the Jacobian matrix of the point-to-cylinder observation equation, which is defined as

$$
\begin{aligned}
\mathbf{H}_{c,i}(\hat{\boldsymbol{x}}_k, \boldsymbol{p}_i^b) &= \frac{\partial h_c(\hat{\boldsymbol{x}}_k \boxplus \delta\boldsymbol{x}_k, \boldsymbol{p}_i^b)}{\partial\delta\boldsymbol{x}_k}\Big|_{\delta\boldsymbol{x}_k=0}, \\
&= [\mathbf{H}_{\delta\boldsymbol{\theta}} \quad \mathbf{H}_{\delta\boldsymbol{p}} \quad \mathbf{0}_{1\times9}]
\end{aligned}
\tag{8}
$$

where $\mathbf{H}_{\delta\boldsymbol{\theta}}$ and $\mathbf{H}_{\delta\boldsymbol{p}}$ represent the Jacobian matrices of the residual w.r.t the attitude error vector and position error vector, respectively, as follows

$$
\begin{aligned}
\mathbf{H}_{\delta\boldsymbol{\theta}} &= -\frac{((\boldsymbol{u}\times)(\mathbf{R}_b^w\boldsymbol{p}_i^b + \boldsymbol{p}^w - \boldsymbol{q}))^T}{||(\boldsymbol{u}\times)(\mathbf{R}_b^w\boldsymbol{p}_i^b + \boldsymbol{p}^w - \boldsymbol{q})||_2}(\boldsymbol{u}\times)\mathbf{R}_b^w(\boldsymbol{p}_i^b\times) \\
\mathbf{H}_{\delta\boldsymbol{p}} &= \frac{((\boldsymbol{u}\times)(\mathbf{R}_b^w\boldsymbol{p}_i^b + \boldsymbol{p}^w - \boldsymbol{q}))^T}{||(\boldsymbol{u}\times)(\mathbf{R}_b^w\boldsymbol{p}_i^b + \boldsymbol{p}^w - \boldsymbol{q})||_2}(\boldsymbol{u}\times)
\end{aligned}
\tag{9}
$$

### 3) Point-to-Plane Measurement Model

The residual of the point $\boldsymbol{p}_i^b$ to the plane can be written as

$$
\begin{aligned}
h_\pi(\boldsymbol{x}_k, \boldsymbol{p}_i^b) &= h_\pi(\hat{\boldsymbol{x}}_k \boxplus \delta\boldsymbol{x}_k, \boldsymbol{p}_i^b) + \boldsymbol{n}_i \\
&\approx h_\pi(\hat{\boldsymbol{x}}_k, \boldsymbol{p}_i^b) + \mathbf{H}_{\pi,i}(\hat{\boldsymbol{x}}_k, \boldsymbol{p}_i^b)\delta\boldsymbol{x}_k + \boldsymbol{n}_i
\end{aligned}
\tag{10}
$$

where $\boldsymbol{n}_i$ represents the Gaussian white noise, and $\mathbf{H}_{\pi,i}$ represents the Jacobian matrix of the point-to-plane observation equation, which is defined as

$$
\begin{aligned}
\mathbf{H}_{\pi,i}(\hat{\boldsymbol{x}}_k, \boldsymbol{p}_i^b) &= \frac{\partial h_\pi(\hat{\boldsymbol{x}}_k \boxplus \delta\boldsymbol{x}_k, \boldsymbol{p}_i^b)}{\partial\delta\boldsymbol{x}_k}\Big|_{\delta\boldsymbol{x}_k=0} \\
&= [-\boldsymbol{u}^T\mathbf{R}_b^w(\boldsymbol{p}_i^b\times) \quad \boldsymbol{u}^T \quad \mathbf{0}_{1\times9}]
\end{aligned}
\tag{11}
$$

where $\boldsymbol{u}$ represents the normal vector of the plane.

Finally, the objective function of the optimization problem can be obtained by combining the prior information and the point-to-plane and point-to-cylinder residuals:

$$
\begin{aligned}
\min_{\delta\boldsymbol{x}_k} \quad & (||\boldsymbol{x}_k \boxminus \hat{\boldsymbol{x}}_k||_{\hat{\mathbf{P}}_k}^2 \\
&+ \sum_{i=1}^{N_\pi} ||h_\pi(\hat{\boldsymbol{x}}_k, \boldsymbol{p}_i^b) + \mathbf{H}_{\pi,i}(\hat{\boldsymbol{x}}_k, \boldsymbol{p}_i^b)\delta\boldsymbol{x}_k||_{\boldsymbol{\Sigma}_\pi}^2 \\
&+ \sum_{i=1}^{N_c} ||h_c(\hat{\boldsymbol{x}}_k, \boldsymbol{p}_i^b) + \mathbf{H}_{c,i}(\hat{\boldsymbol{x}}_k, \boldsymbol{p}_i^b)\delta\boldsymbol{x}_k||_{\boldsymbol{\Sigma}_c}^2)
\end{aligned}
\tag{12}
$$

where $N_\pi$ and $N_c$ represent the number of point-to-plane and point-to-cylinder observations, respectively, and $\hat{\mathbf{P}}_k$, $\boldsymbol{\Sigma}_\pi$ and $\boldsymbol{\Sigma}_c$ represent the observation noise covariance matrices of the prior information, the point-to-plane, and point-to-cylinder observations, respectively. We use the residual covariance matrix of the cylinder fitting as $\boldsymbol{\Sigma}_c$.

## IV. EXPERIMENTS AND RESULTS

### A. Implementation and Evaluation Setup

The proposed SE-LIO is implemented in C++ based on the robot operating system (ROS). Field tests are conducted using a low-speed wheeled robot with an average speed of around 1.5 m/s. The system uses a solid-state LiDAR with a frame rate of 10 Hz (Livox Mid-70), an industrial-grade MEMS IMU (ADI ADIS16465 with a gyroscope bias instability of 2 °/hr and a frame rate of 200 Hz), and a dual-antenna GNSS receiver with a frame rate of 1 Hz (NovAtel OEM-718D). The GNSS real-time kinematic (RTK) technique is adopted to achieve high-accuracy positioning. All sensors are synchronized through hardware triggers to the GNSS time. The ground-truth system is a high-accuracy GNSS/INS integrated navigation system using the GNSS-RTK and a navigation-grade IMU. The ground truth (with an accuracy of 0.02 m for position and 0.01° for attitude) is generated by post-processing software.

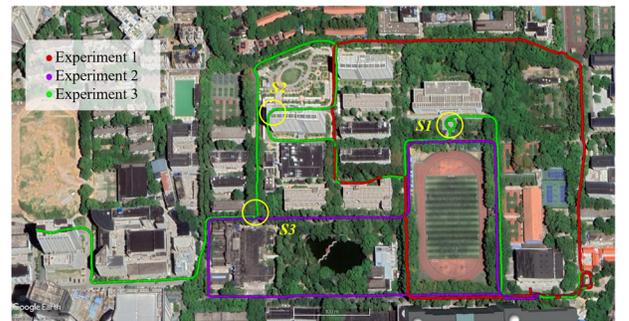

(a) Experiments 1, 2, and 3

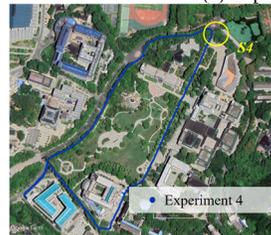

(b) Experiment 4

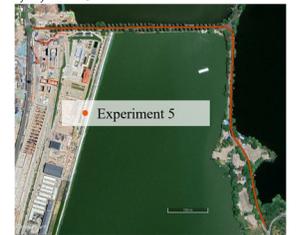

(c) Experiment 5

Fig. 5. Test environments for quantitative evaluation. *S1*, *S2*, and *S3* represent the degraded scenes in Experiment 2, and *S4* represents the degraded scene in Experiment 4.

The deep learning model utilized for segmentation requires retraining for different types of LiDAR. Currently, there is no suitable public dataset containing solid-state LiDAR and tree-rich test scenes; therefore, the system was not tested on public datasets. To quantitatively evaluate the accuracy of the proposed system, we conducted a series of field tests in different environments. The test environments include campus and park areas, as shown in Fig. 5. The experiments, numbered 1 through 4, were conducted at Wuhan University, where the environment featured numerous trees and buildings. The last experiment, numbered 5, was conducted in Donghu, Wuhan, a large lake with artificial roads and trees on both sides. All the test environments contain many dynamic objects. *S1~S3* in Fig. 5 represent the degraded scenes of Experiment 2, and *S4* represents the degraded scene of Experiment 4.

To underscore the merits of the proposed method, we conducted a comparative analysis with the LIO system that solely utilizes plane features. The proposed method is referred to as SE-LIO. The baseline method is referred to as Ori-LIO, which only uses plane features. The performance of Ori-LIO is comparable to that of FAST-LIO, but it has been further optimized, as mentioned in Section III-C. Therefore, we use Ori-LIO as the baseline algorithm. To verify the effectiveness of the cylindrical features, we also conducted ablation experiments, *i.e.*, testing the positioning accuracy that solely removed the unstructured point clouds and compared it with the proposed SE-LIO. The method that solely removes unstructured point clouds is referred to as SE-LIO-RU, where RU stands for removing unstructured point clouds.

The positioning performance was evaluated based on absolute and relative pose errors. It is important to note that the results are deterministic in each run. The system was run in real-time on a desktop PC (Intel Core i7-11700 CPU @ 2.50 GHz, 32 GB RAM, and an NVIDIA GTX 1650 GPU) under the ROS framework. It should be noted that different data used the same parameters for the same tested method.

### B. The Impact of Point-cloud Integration on Semantic Segmentation Accuracy

We used multi-frame merging to achieve point-cloud integration. The LiDAR point cloud was motion-compensated using the INS pose during multi-frame merging. The duration of a single frame of point cloud was 0.1 s, and the number of merged frames ranged from 1 to 6, corresponding to a merging time of 0.1 to 0.6 s. The test results are shown in TABLE I. Here, f represents frame(s), *i.e.*, the number of frames.

As shown in TABLE I, as the number of merged frames increases, the semantic segmentation accuracy also increases. The results indicate that improving the point-cloud coverage is helpful for semantic segmentation. However, semantic segmentation accuracy is almost unchanged when the merged frames exceed 3. Improving the number of merged frames can reduce the frequency of semantic segmentation, but it will increase the error of INS motion compensation. Therefore, to balance the frequency of semantic segmentation and the accuracy of INS motion compensation, we merged 5 frames of point clouds for semantic segmentation in the subsequent tests.

TABLE I
SEMANTIC SEGMENTATION ACCURACY UNDER DIFFERENT MERGING FRAMES

| class | Class IOUs | | | | | |
|---|---|---|---|---|---|---|
| | 1f | 2f | 3f | 4f | 5f | 6f |
| *car* | 0.51 | 0.76 | 0.767 | 0.77 | 0.75 | **0.78** |
| *motor* | 0.11 | 0.44 | 0.56 | 0.54 | 0.55 | **0.61** |
| *road* | 0.93 | 0.96 | 0.968 | **0.97** | **0.97** | 0.97 |
| *building* | 0.23 | 0.39 | 0.40 | 0.41 | 0.40 | **0.43** |
| *vegetation* | 0.81 | 0.85 | 0.85 | 0.85 | 0.85 | **0.86** |
| *trunk* | 0.51 | 0.62 | 0.63 | 0.63 | 0.64 | **0.67** |
| MEAN | 0.52 | 0.67 | 0.70 | 0.70 | 0.69 | **0.72** |

TABLE II
ARE AND ATE OF THE THREE METHODS

| ARE/ATE (deg/m) | Ori-LIO | SE-LIO-RU | SE-LIO |
|---|---|---|---|
| *E1* | 0.858 / 2.586 | 0.680 / 2.339 | **0.547 / 1.850** |
| *E2* | 0.619 / 1.252 | **0.282** / 0.890 | 0.344 / **0.746** |
| *E3* | 0.526 / 1.252 | 0.515 / 1.322 | **0.413 / 0.936** |
| *E4* | 0.563 / 1.207 | **0.250** / 0.915 | 0.285 / **0.634** |
| *E5* | 0.328 / 1.370 | 0.214 / 0.666 | **0.200 / 0.199** |
| MEAN | 0.579 / 1.533 | 0.388 / 1.226 | **0.358 / 0.873** |

### C. Evaluation of the Positioning Accuracy

The proposed method employs an adaptive piecewise fitting cylinder model, necessitating the configuration of the max fitting depth. This section sets the max fitting depth to 3, with a comprehensive explanation to follow in Section IV-D. The absolute rotation error (ARE) and absolute translation error (ATE) are shown in TABLE II, with the superior result among the three emphasized in bold, and E1~E5 represent Experiment 1~Experiment 5, respectively.

As depicted in TABLE II, SE-LIO-RU exhibits a substantial enhancement over Ori-LIO in most tests, and SE-LIO exhibits superior performance to SE-LIO-RU in both attitude and position accuracy. Specifically, in Experiment 5, the test scene is situated by a lake. During the majority of Experiment 5, the point cloud in the horizontal direction is only unstructured point clouds, such as trees and dynamic objects, and lacks structured point clouds, such as buildings. Therefore, the absolute positioning accuracy is greatly improved after removing unstructured point clouds. Furthermore, the proposed method utilizes pole-like objects like tree trunks for positioning, thereby strengthening the horizontal constraint and enhancing the absolute positioning accuracy compared to SE-LIO-RU.

Fig. 6 shows the trajectories estimated by the three methods in Experiment 5. SE-LIO exhibits the least drift, aligning more closely with the ground truth. In contrast, both SE-LIO-RU and Ori-LIO demonstrate larger drift. The relative translation error (RTE) and relative rotation error (RRE) were also evaluated to provide insight into short-term accuracy. The results are presented in TABLE III, with the superior result among the three emphasized in bold. As exhibited in TABLE III, the proposed SE-LIO outperforms the other two methods in most tests. The short-term accuracy of the proposed SE-LIO is improved, reflecting the system's superior robustness.

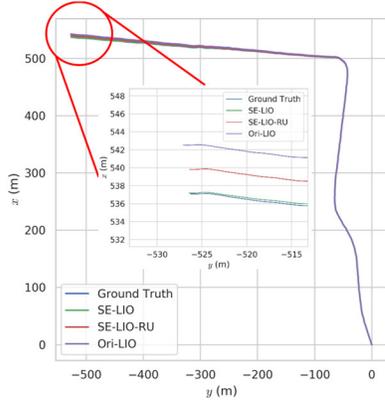

Fig. 6. Trajectories estimated by the three methods in Experiment 5.

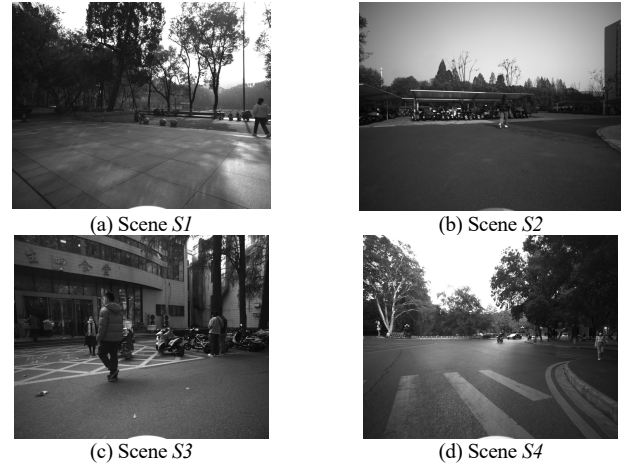

(a) Scene *S1*    (b) Scene *S2*

(c) Scene *S3*    (d) Scene *S4*

Fig. 7. Scenes selected for evaluating the robustness of the proposed method in tree-rich environments.

TABLE III
RRE AND RTE OF THE THREE METHODS

| RRE/RT E (deg/%) | 25m | | | 100m | | |
|---|---|---|---|---|---|---|
| | Ori-LIO | SE-LIO-RU | SE-LIO | Ori-LIO | SE-LIO-RU | SE-LIO |
| *E1* | 0.31/**0.86** | 0.30/0.89 | **0.30**/0.87 | 0.46/0.62 | 0.37/0.61 | **0.34/0.61** |
| *E2* | 0.15/0.87 | 0.10/0.87 | **0.10/0.82** | 0.30/**0.61** | 0.17/0.64 | **0.15**/0.63 |
| *E3* | 0.20/1.01 | 0.16/0.98 | **0.15/0.92** | 0.32/0.72 | 0.29/0.70 | **0.24/0.67** |
| *E4* | 0.16/0.84 | 0.13/0.85 | **0.12/0.77** | 0.27/0.61 | 0.17/0.57 | **0.16/0.53** |
| *E5* | 0.10/0.48 | 0.10/0.46 | **0.09/0.43** | 0.18/0.41 | 0.14/**0.32** | **0.10**/0.34 |
| MEAN | 0.18/0.81 | 0.16/0.81 | **0.15/0.76** | 0.31/0.59 | 0.23/0.57 | **0.20/0.56** |

TABLE IV
ARE AND ATE OF THE METHODS WITH DIFFERENT MAX DEPTH

| ARE/ATE (deg/m) | D1 | D2 | D3 | D4 |
|---|---|---|---|---|
| *E1* | 0.579/1.908 | 0.576/1.877 | **0.547/1.850** | 0.646/2.209 |
| *E2* | 0.340/0.794 | **0.333/0.735** | 0.344/0.746 | 0.340/0.757 |
| *E3* | 0.480/1.165 | 0.460/1.091 | **0.413/0.936** | 0.446/1.040 |
| *E4* | 0.272/0.640 | **0.266/0.569** | 0.285/0.634 | 0.280/0.667 |
| *E5* | 0.205/**0.186** | 0.201/0.196 | **0.200**/0.199 | 0.202/0.199 |
| *MEAN* | 0.375/0.939 | 0.367/0.894 | **0.358/0.873** | 0.383/0.974 |

### D. The Impact of Adaptive Piecewise Cylinder Fitting on Positioning Accuracy

As previously discussed, we employ a binary tree to implement adaptive piecewise fitting. Different piecewise fitting results can be obtained by adjusting the binary tree's max depth. When the max depth is set to $i$, the point cloud of this pole-like objects cluster will be divided into $2^{i-1}$ segments at most. In the tests above, the impact of piecewise cylinder fitting on positioning accuracy was evaluated by setting the max depth of the binary tree to 1, 2, 3, and 4, respectively. A max depth of 1 indicates no piecewise fitting, and the results of different piecewise fitting depths are denoted as D1, D2, D3, and D4, respectively. The absolute pose error is shown in TABLE IV, where the best result among the four is highlighted in bold.

As can be seen from TABLE IV, when the max depth of the binary tree is 1, the positioning accuracy of the three tests is suboptimal. In Experiments 1, 2, 3, and 4, positioning accuracy is improved when the max depth is either 2 or 3, compared to a max depth of 1. However, increasing the max depth to 4 decreases positioning accuracy. These results indicate that piecewise fitting has a certain impact on positioning accuracy, with optimal positioning accuracy achieved at a max depth of

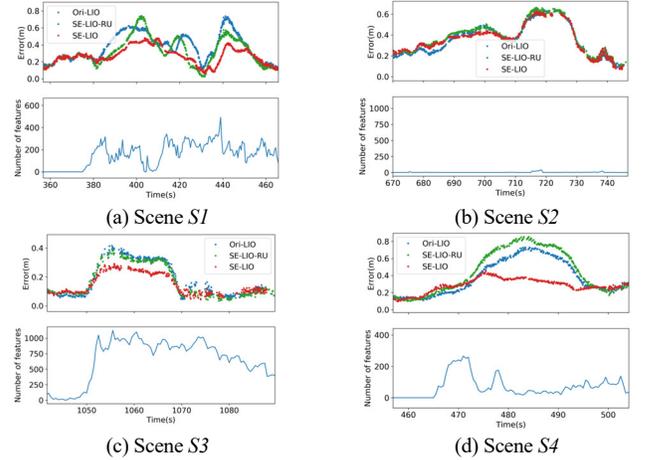

(a) Scene *S1*    (b) Scene *S2*

(c) Scene *S3*    (d) Scene *S4*

Fig. 8. RTEs curve of the robot over a distance of 25 m in Scenes *S1~S4*.

3. Further increases in max depth do not enhance accuracy but increase computational complexity. Consequently, the max depth of the binary tree is set to 3.

### E. Evaluation of the Robustness

We selected several representative scenes to evaluate the robustness of the proposed method in tree-rich environments, as depicted in Fig. 5. Scenes *S1* to *S3* represent the degraded conditions of Experiment 2, while Scene *S4* corresponds to the degraded condition of Experiment 4. The robot performed a turning maneuver in all scenes. Pole-like object features are lacking in Scene *S2* but are observable in Scenes *S1*, *S3*, and *S4*. The corresponding scenes of *S1~S4* are shown in Fig. 7.

The RTEs curve of the robot over a distance of 25 m in these scenes is depicted in Fig. 8. In each subfigure, the upper part represents the RTEs of the robot, and the lower part represents the number of cylindrical features observed by the robot. As inferred from Fig. 8, in Scenes *S1*, *S3*, and *S4*, SE-LIO demonstrates superior positioning accuracy compared to Ori-LIO and SE-LIO-RU. It suggests that the proposed method exhibits robustness in environments rich in trees. The cylindrical features significantly enhance the robot's positioning accuracy during turning maneuvers by providing additional horizontal constraint information, compensating for the limitations of plane features during such maneuvers.

Notably, in Scene *S2*, the positioning accuracy of SE-LIO is comparable to that of Ori-LIO and SE-LIO-RU, indicating that the proposed method does not compromise accuracy in scenes devoid of pole-like object features.

In Scene *S4*, the positioning accuracy of SE-LIO-RU is notably the least optimal. It can be attributed to the fact that in Scene *S4*, the LiDAR scans fewer plane features, such as buildings. Ori-LIO makes use of unstructured point clouds while SE-LIO-RU removes them. Despite the unreliability of the plane features extracted from unstructured point clouds, they provide horizontal constraints, resulting in better positioning accuracy than SE-LIO-RU. However, SE-LIO, which is based on SE-LIO-RU, utilizes cylindrical features that provide more reliable horizontal constraints. Consequently, SE-LIO demonstrates superior positioning accuracy compared to both Ori-LIO and SE-LIO-RU.

*F. Runtime Analysis*

We further tested the runtime of the proposed method. The test environment comprised an Intel Core i7-11700 CPU @ 2.50 GHz, 32 GB RAM, and an NVIDIA GTX 1650 GPU. In Section IV-B, we merged 5 frames of point clouds (0.5 s) for semantic segmentation and LIO positioning. For a 0.5 s point cloud, the average runtime for semantic segmentation was 392.8 ms, while the LIO fusion took 33.5 ms. Semantic segmentation accounted for approximately 90% of the total runtime, indicating potential for optimization in future work.

## V. Conclusion

In this study, we propose a semantic-enhanced solid-state-LIO method for environments abundant with trees, such as campuses and parks. The method integrates and compensates multiple LiDAR frames using the INS pose to address the problem of low semantic segmentation accuracy due to sparse point clouds. Semantic information is then employed to enhance the positioning performance, including removing unstructured point clouds and constructing point-to-cylinder constraints using the cylindrical features of pole-like objects. An adaptive piecewise fitting method is proposed that the pole-like object is segmented into multiple cylinders, obtaining more accurate cylindrical features. Hence, the positioning accuracy can be improved in scenes with curved tree trunks. Experimental results demonstrate that SE-LIO outperforms the baseline method in terms of positioning accuracy and robustness.

## References

[1] P. Biber and W. Strasser, "The normal distributions transform: a new approach to laser scan matching," in *Proceedings 2003 IEEE/RSJ International Conference on Intelligent Robots and Systems (IROS 2003) (Cat. No.03CH37453)*, Las Vegas, Nevada, USA, 2003, vol. 3, pp. 2743–2748.

[2] M. Magnusson, A. Lilienthal, and T. Duckett, "Scan registration for autonomous mining vehicles using 3D-NDT," *J. Field Robot.*, vol. 24, no. 10, pp. 803–827, 2007,

[3] C. Yuan and X. Liu, "Efficient and Probabilistic Adaptive Voxel Mapping for Accurate Online LiDAR Odometry," *IEEE Robot. Autom. Lett.*, vol. 7, no. 3, 2022.

[4] T. Shan and B. Englot, "LeGO-LOAM: Lightweight and Ground-Optimized Lidar Odometry and Mapping on Variable Terrain," in *2018 IEEE/RSJ International Conference on Intelligent Robots and Systems (IROS)*, Oct. 2018, pp. 4758–4765.

[5] I. Bogoslavskyi and C. Stachniss, "Fast range image-based segmentation of sparse 3D laser scans for online operation," in *2016 IEEE/RSJ International Conference on Intelligent Robots and Systems (IROS)*, Daejeon, South Korea, Oct. 2016, pp. 163–169.

[6] Y. Guo, H. Wang, Q. Hu, H. Liu, L. Liu, and M. Bennamoun, "Deep Learning for 3D Point Clouds: A Survey," *IEEE Trans. Pattern Anal. Mach. Intell.*, vol. 43, no. 12, pp. 4338–4364, Dec. 2021,

[7] B. Wu, A. Wan, X. Yue, and K. Keutzer, "SqueezeSeg: Convolutional Neural Nets with Recurrent CRF for Real-Time Road-Object Segmentation from 3D LiDAR Point Cloud," in *2018 IEEE International Conference on Robotics and Automation (ICRA)*, Brisbane, QLD, May 2018, pp. 1887–1893.

[8] A. Milioto, I. Vizzo, J. Behley, and C. Stachniss, "RangeNet ++: Fast and Accurate LiDAR Semantic Segmentation," in *2019 IEEE/RSJ International Conference on Intelligent Robots and Systems (IROS)*, Macau, China, Nov. 2019, pp. 4213–4220.

[9] R. A. Rosu, P. Schütt, J. Quenzel, and S. Behnke, "LatticeNet: fast spatio-temporal point cloud segmentation using permutohedral lattices," *Auton. Robots*, vol. 46, no. 1, pp. 45–60, Jan. 2022,

[10] D. Rethage, J. Wald, J. Sturm, N. Navab, and F. Tombari, "Fully-Convolutional Point Networks for Large-Scale Point Clouds," in *Computer Vision – ECCV 2018*, vol. 11208, V. Ferrari, M. Hebert, C. Sminchisescu, and Y. Weiss, Eds. Cham: Springer International Publishing, 2018, pp. 625–640.

[11] C. R. Qi, L. Yi, H. Su, and L. J. Guibas, "PointNet++: Deep Hierarchical Feature Learning on Point Sets in a Metric Space," in *Advances in Neural Information Processing Systems*, 2017, vol. 30.

[12] Q. Hu *et al.*, "RandLA-Net: Efficient Semantic Segmentation of Large-Scale Point Clouds," in *2020 IEEE/CVF Conference on Computer Vision and Pattern Recognition (CVPR)*, Seattle, WA, USA, Jun. 2020, pp. 11105–11114.

[13] X. Chen, A. Milioto, E. Palazzolo, P. Giguère, J. Behley, and C. Stachniss, "SuMa++: Efficient LiDAR-based Semantic SLAM." May 24, 2021.

[14] G. Chen, B. Wang, X. Wang, H. Deng, B. Wang, and S. Zhang, "PSF-LO: Parameterized Semantic Features Based Lidar Odometry," in *2021 IEEE International Conference on Robotics and Automation (ICRA)*, Xi'an, China, May 2021, pp. 5056–5062.

[15] S. W. Chen *et al.*, "SLOAM: Semantic Lidar Odometry and Mapping for Forest Inventory," *IEEE Robot. Autom. Lett.*, vol. 5, no. 2, pp. 612–619, Apr. 2020,

[16] L. Li *et al.*, "SA-LOAM: Semantic-aided LiDAR SLAM with Loop Closure," in *2021 IEEE International Conference on Robotics and Automation (ICRA)*, May 2021, pp. 7627–7634.

[17] J. Lin and F. Zhang, "Loam livox: A fast, robust, high-precision LiDAR odometry and mapping package for LiDARs of small FoV," in *2020 IEEE International Conference on Robotics and Automation (ICRA)*, May 2020, pp. 3126–3131.

[18] H. Tang, X. Niu, T. Zhang, L. Wang, and J. Liu, "LE-VINS: A Robust Solid-State-LiDAR-Enhanced Visual-Inertial Navigation System for Low-Speed Robots," *IEEE Trans. Instrum. Meas.*, vol. 72, pp. 1–13, 2023,

[19] J. Zhang and S. Singh, "LOAM: Lidar Odometry and Mapping in Real-time," in *Robotics: Science and Systems X*, Jul. 2014.

[20] T. Shan, B. Englot, D. Meyers, W. Wang, C. Ratti, and D. Rus, "LIO-SAM: Tightly-coupled Lidar Inertial Odometry via Smoothing and Mapping," in *2020 IEEE/RSJ International Conference on Intelligent Robots and Systems (IROS)*, Las Vegas, NV, USA, Oct. 2020, pp. 5135–5142.

[21] W. Xu and F. Zhang, "FAST-LIO: A Fast, Robust LiDAR-Inertial Odometry Package by Tightly-Coupled Iterated Kalman Filter," *IEEE Robot. Autom. Lett.*, vol. 6, no. 2, pp. 3317–3324, Apr. 2021,

[22] W. Xu, Y. Cai, D. He, J. Lin, and F. Zhang, "FAST-LIO2: Fast Direct LiDAR-Inertial Odometry," *IEEE Trans. Robot.*, vol. 38, no. 4, pp. 2053–2073, Aug. 2022,

[23] R. Schnabel, R. Wahl, and R. Klein, "Efficient RANSAC for Point-Cloud Shape Detection," *Comput. Graph. Forum*, vol. 26, no. 2, pp. 214–226, 2007,

[24] K. Khan, S. U. Rehman, K. Aziz, S. Fong, and S. Sarasvady, "DBSCAN: Past, present and future," in *The Fifth International Conference on the Applications of Digital Information and Web Technologies (ICADIWT 2014)*, Feb. 2014, pp. 232–238.

[25] E.-H. Shin, "Estimation techniques for low-cost inertial navigation," *UCGE Rep.*, vol. 20219, 2005.